\documentclass[journal,twoside,web]{ieeecolor}
\pdfoutput=1
\usepackage{etoolbox}
\usepackage{tmi}
\usepackage{cite}
\usepackage{amsmath,amssymb,amsfonts}
\usepackage{algorithmic}
\usepackage{graphicx}
\usepackage{textcomp}
\usepackage[switch]{lineno}
\usepackage{hyperref}
\modulolinenumbers[5]
\usepackage{bm}
\usepackage{multirow}
\usepackage{booktabs}
\usepackage{enumerate}
\usepackage{listings}
\usepackage{diagbox}
\lstdefinestyle{code}{
  language=Python,
  backgroundcolor=\color[rgb]{1.0, 1.0, 1.0}, 
  commentstyle=\color[rgb]{0, 0.6, 0},
  keywordstyle=\color{magenta},
  numberstyle=\tiny\color[rgb]{0.5, 0.5, 0.5},
  stringstyle=\color[rgb]{0.58, 0, 0.82},
  basicstyle=\ttfamily\footnotesize,
  breakatwhitespace=false,
  breaklines=true,
  captionpos=b,
  keepspaces=true,
  numbers=none,
  numbersep=4pt,
  showspaces=false,
  showstringspaces=false,
  showtabs=false,
  tabsize=2
}
\usepackage{color}
\definecolor{highlight}{rgb}{0,0.54,0.86} 

\newcommand{\figref}[1]{Fig.~\ref{#1}}
\newcommand{\tabref}[1]{Table~\ref{#1}}
\newcommand{\eqtref}[1]{Eq.~\ref{#1}}

\def\BibTeX{{\rm B\kern-.05em{\sc i\kern-.025em b}\kern-.08em
    T\kern-.1667em\lower.7ex\hbox{E}\kern-.125emX}}
\begin{document}
\title{Rethinking Attention-Based Multiple Instance Learning for Whole-Slide Pathological Image Classification: An Instance Attribute Viewpoint}
\author{Linghan Cai, Shenjin Huang, Ye Zhang, Jinpeng Lu, and Yongbing Zhang, \IEEEmembership{Senior Member, IEEE}
\thanks{This work was supported in part by the National Natural Science Foundation of China under 62031023 \& 62331011; in part by the Shenzhen Science and Technology Project under JCYJ20200109142808034 \& GXWD20220818170353009, and in part by the Fundamental Research Funds for the Central Universities under HIT.OCEF.2023050. Thanks to the doctors from Haiyu Zhou's Group at Guangdong Provincial People's Hospital for providing medical knowledge support for this work.
\textit{Corresponding author: Yongbing Zhang.}}
\thanks{LH Cai, Y Zhang, and YB Zhang are with the School of Computer Science and Technology, Harbin Institute of Technology, Shenzhen, 518055, China. (e-mails: cailh@stu.hit.edu.cn; zhangye94@stu.hit.edu.cn; ybzhang08@hit.edu.cn).}
\thanks{SJ Huang is with the Faculty of Computing, Harbin Institute of Technology, Harbin, 150001, China. (e-mail: shenjinhuang@stu.hit.edu.cn).}
\thanks{JP Lu is with the School of Science, Harbin Institute of Technology, Shenzhen, 518055, China. (e-mail: jinpenglu231@gmail.com).}}
\maketitle

\begin{abstract}
Multiple instance learning (MIL) is a robust paradigm for whole-slide pathological image (WSI) analysis, processing gigapixel-resolution images with slide-level labels. As pioneering efforts, attention-based MIL (ABMIL) and its variants are increasingly becoming popular due to the characteristics of simultaneously handling clinical diagnosis and tumor localization. However, the attention mechanism exhibits limitations in discriminating between instances, which often misclassifies tissues and potentially impairs MIL performance. This paper proposes an Attribute-Driven MIL (AttriMIL) framework to address these issues. Concretely, we dissect the calculation process of ABMIL and present an attribute scoring mechanism that measures the contribution of each instance to bag prediction effectively, quantifying instance attributes. Based on attribute quantification, we develop a spatial attribute constraint and an attribute ranking constraint to model instance correlations within and across slides, respectively. These constraints encourage the network to capture the spatial correlation and semantic similarity of instances, improving the ability of AttriMIL to distinguish tissue types and identify challenging instances. Additionally, AttriMIL employs a histopathology adaptive backbone that maximizes the pre-trained model's feature extraction capability for collecting pathological features. Extensive experiments on three public benchmarks demonstrate that our AttriMIL outperforms existing state-of-the-art frameworks across multiple evaluation metrics. The implementation code is available at \href{https://github.com/MedCAI/AttriMIL}{https://github.com/MedCAI/AttriMIL}.
\end{abstract}


\section{Introduction}
\label{sec:introduction}
\IEEEPARstart{H}{istopathological} examination is regarded as the gold standard for cancer diagnosis and prognosis in modern healthcare \cite{lu2021ai}. During these examinations, pathologists use a microscope to view specimens on stained slides, identifying tumor areas within tissue slices. With advances in scanning technology, traditional glass slides are increasingly being converted into digital whole-slide images (WSIs), providing significant opportunities for computer-assisted diagnosis. Nevertheless, the application of computational techniques in histopathological image analysis encounters great challenges: the enormous resolution of WSIs (e.g., 40,000$\times$40,000 pixels) and the lack of fine-grained (patch-level) annotations \cite{srinidhi2021deep}. Consequently, WSI classification is often formulated as a weakly supervised task using multiple instance learning (MIL) \cite{campanella2019clinical, qu2022dgmil, zhang2022dtfd}. In the MIL paradigm, each WSI is treated as a labeled bag containing multiple unlabeled instances (tile patches). If a bag is negative, all instances within it are negative. Conversely, a positive WSI contains at least one positive instance.

\begin{figure}[!t]
\centerline{\includegraphics[width=0.95\columnwidth]
{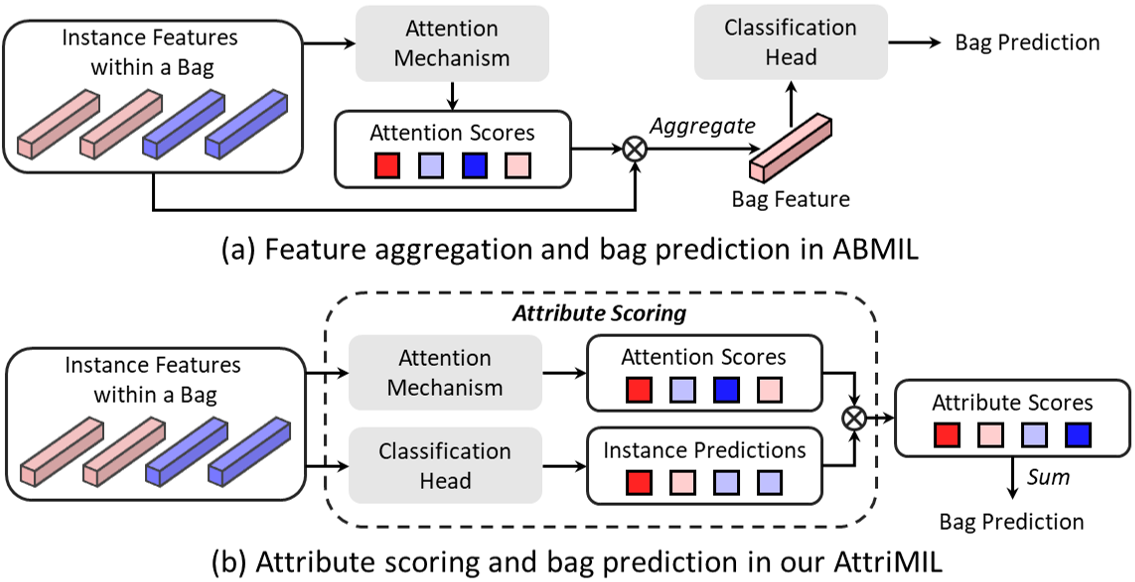}}
\caption{Illustration of ABMIL and our AttriMIL. For features, red cuboids denote positive attributes and blue cuboids are negative; for scores, redder colors represent higher scores and bluer colors indicate lower ones. Notably, for a positive bag, existing methods \cite{abmil, clam, wang2023hard, qu2022dgmil, zhang2022dtfd} generally believe the instances with high attention scores are positive instances.}
\label{fig:Scoring}
\end{figure}

In general, WSI classification requires the MIL framework to complete two tasks: bag classification and instance discrimination, corresponding to clinical diagnosis and tumor localization, respectively. As a seminal work, attention-based MIL (ABMIL) \cite{abmil} addresses both tasks simultaneously and has therefore been widely adopted in WSI analysis. This architecture processes the input WSI in two steps: embedding image patches as instance-level features, and then aggregating the features into a bag-level representation. In the aggregation phase, an attention pooling operation assigns each instance an attention score, which weights the contribution to the bag aggregation and, to a certain extent, reflects the attribute of the instance (shown in \figref{fig:Scoring} (a)). Building on the attention mechanism, subsequent studies \cite{clam, wang2023hard, shi2023structure, PMIL} treat high-attention patches within a positive WSI as positive instances and exploit contrastive learning or prototype learning methods to refine the instance representation for improving classification accuracy.

However, an open question remains in ABMIL: Can attention reliably distinguish between instances? This issue is critical as introducing mislabeled instances could harm MIL and confuse diagnosis \cite{qu2023rethinking}. In this paper, we argue that the level of attention in ABMIL is not a reliable measure of instance attributes for two reasons. Firstly, both positive and negative patches contribute to bag prediction. Typical negative instances have high attention, offering high confidence for negative bag prediction and prompting the network to identify positive instances. Secondly, a high level of attention does not necessarily indicate a high contribution to the prediction. As shown in \figref{fig:Scoring} (a), the outcome of ABMIL is determined by both the attention pooling and bag classification head. Hence, instance discrimination based merely on attention level is incomplete and potentially misleading, and thus an effective measure of instance attributes is worth exploring.

Besides the aforementioned issue, the assumption in MIL algorithms \cite{abmil, chen2006miles, carbonneau2018multiple} that instances are independent limits WSI classification performance due to the negligence of significant correlations between patches. The correlations can be categorized into two types within this paper: intra-slide correlation and inter-slide correlation. For example, considering a tumor area, the intra-slide correlation emphasizes the spatial distribution of instances within a WSI, reflecting how tumor cells tend to cluster \cite{cheung2016collective}. Similarly, the inter-slide relation reveals attribute consistency among tumor instances of the same subtype \cite{travis2002pathology}. To model instance correlations, recent works \cite{chen2022scaling, transmil, li2023survival} investigate Transformer \cite{dosovitskiy2020image} for aggregating instances. Despite great progress, the use of position encoding hinders the flexibility of MIL frameworks, as it imposes rigid spatial relationships. Furthermore, these methods do not establish instance connections across WSIs, leading to suboptimal generalization and difficulty in identifying hard instances.

Motivated by the above discussions, we present an Attribute-Driven Multiple Instance Learning (AttriMIL) that comprehensively improves ABMIL in the pathological image classification task. To precisely represent instance attributes, we introduce an attribute scoring mechanism (shown in \figref{fig:Scoring} (b)) that integrates attention pooling with bag classification head to quantify instances' contribution to bag prediction. Based on attribute scores, a spatial attribute constraint is applied to maintain the spatial correlations among instances within a single WSI. Meanwhile, we develop an attribute ranking loss to model the instance correlation across WSIs, enhancing the network's capability to distinguish challenging instances by emphasizing differences between positive and negative instances. Moreover, to model these correlations effectively, we employ a histopathology adaptive backbone that optimizes the pre-trained model at different stages, maximizing the model's pathological feature extraction ability. In summary, the main contributions of this paper are as follows:
\begin{itemize}
    \item An attribute-driven multiple instance learning (AttriMIL) framework is proposed for histopathological image analysis. In detail, AttriMIL improves the instance aggregation and bag prediction process of attention-based multiple instance learning (ABMIL) through attribute scoring, achieving effective instance attribute measurement.
    \item Considering patch correlations in WSIs, we introduce two instance constraints for regularizing the training of the MIL framework. For a WSI, a spatial attribute constraint is leveraged to model the spatial dependency between instances. For a set of WSIs, an attribute ranking loss is designed to highlight the attribute differences between instances of different subtypes.
    \item Inspired by the parameter-efficient fine-tuning technique \cite{he2023parameter}, we develop a histopathology adaptive backbone for efficient pathological feature extraction, where the backbone enables our AttriMIL to model instance correlations at multiple feature levels.
    \item Extensive experiments validate the effectiveness of the proposed approaches, and AttriMIL derives state-of-the-art results on three public datasets. Notably, AttriMIL also exhibits potential in identifying out-of-detection (OOD) samples, offering a promising solution for developing a complete computer-aided pathological diagnosis system.
\end{itemize}

The rest of this study is organized as follows. Section \ref{sec:related} reviews MIL and parameter-efficient fine-tuning. Our method is described in Section \ref{sec:methodology}. Section \ref{sec:experiments} provides experimental results. Discussion and conclusion are given in Section \ref{sec:disandcon}.

\begin{figure*}[t]
\centerline{\includegraphics[width=0.95\textwidth]
{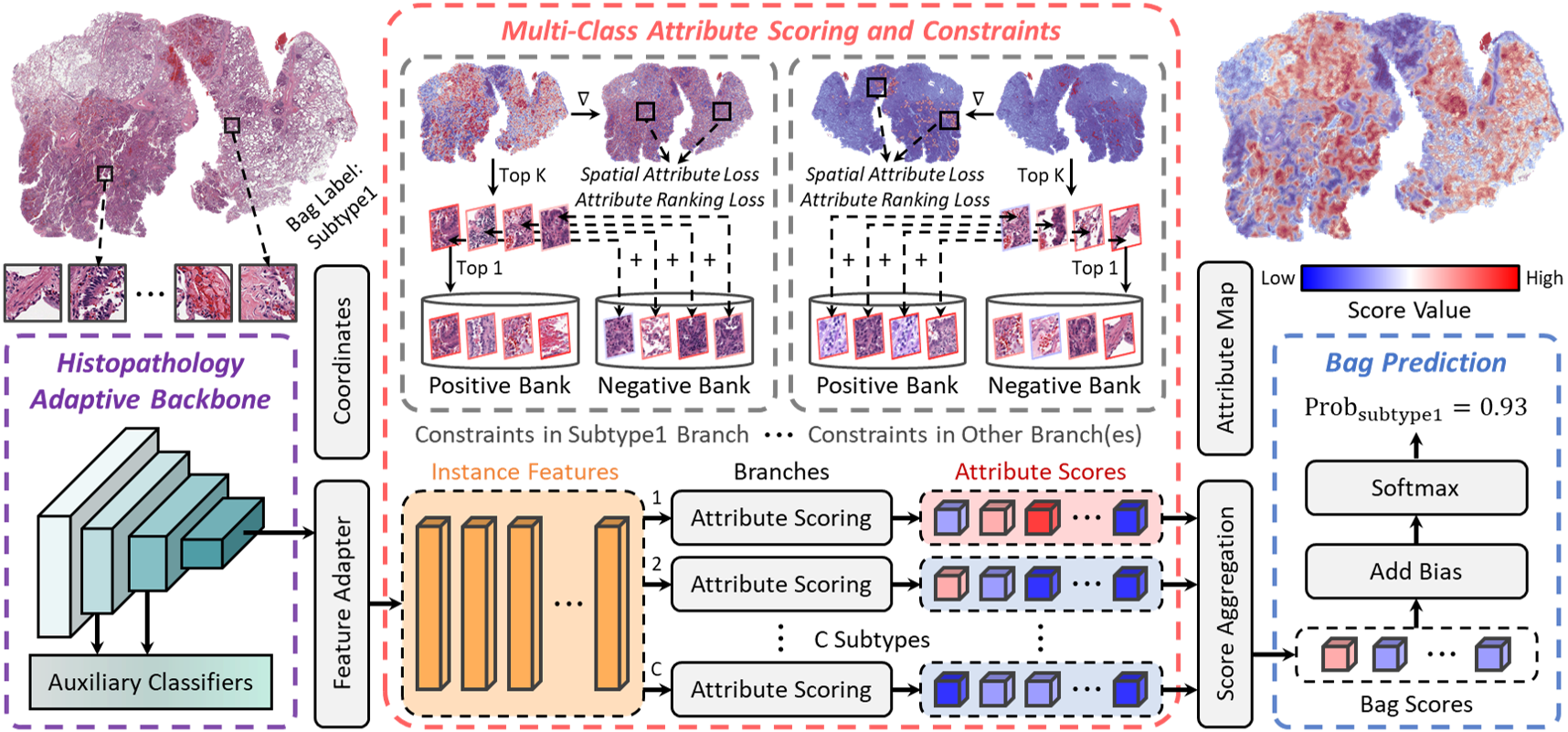}}
\caption{Overview of AttriMIL framework. For an input WSI, AttriMIL crops it into patches and adopts a histopathology adaptive backbone to extract instance features. Afterward, AttriMIL generates instance attribute scores in each subtype branch (tumor and normal in the tumor detection task) using a multi-class attribute scoring mechanism. For a subtype branch, it considers WSIs of the same subtype as it as positive and WSIs of the other subtypes as negative. Spatial attribute constraint (``$\nabla$'' is a differential operation) and attribute ranking constraint (``$+$'' denotes a weighted sum operation) are applied in the training stage. Next, AttriMIL performs score aggregation to obtain $\text{C}$ bag scores and then generates bag prediction probabilities. Instance attribute scores corresponding to the bag prediction are mapped for tumor localization.}
\label{fig:overview}
\end{figure*}

\section{Related Work}
\label{sec:related}
\subsection{Instance-based MIL on WSIs}
MIL methods can be broadly divided into two categories in WSI analysis, namely, instance-based MIL and bag-based MIL. The main idea of instance-based MIL methods is to train an instance classifier and then collect the instance predictions for bag prediction. Early solutions \cite{kohavi1997wrappers} adopt a straightforward MIL framework that propagates the bag label to instances for instance classifier's training. This behavior inevitably brings noisy instance-level supervision due to the presence of only a small portion of positive regions within a WSI \cite{lin2023interventional}. To mitigate this problem, several studies \cite{tourniaire2023ms} explore a mixed-supervision strategy, which utilizes patch-level annotations for partial instances, effectively reducing the adverse effects of noise by focusing on the labeled instances. Without instance-level labels, Hou \textit{et al.} \cite{hou2016patch} propose a PatchCNN that uses a subtle threshold scheme to select representative instances at both class and WSI levels. Lin \textit{et al.} \cite{lin2022interventional} summarize the previous studies from a causal perspective and develop an IMIL to choose key instances via causal intervention and effects. Albeit the fruitful progress, the performance of instance-based MIL is usually inferior to bag-based methods in WSI classification due to the inaccurate instance-level supervision signal.

\subsection{Bag-based MIL on WSIs}
Bag-based MIL aggregates instance features into bag representation and classifies them through distance measures or bag classifiers \cite{srinidhi2021deep}. In WSI classification, executing a bag-based MIL is non-trivial owing to the large memory required to store patches. Therefore, current methods tend to adopt a two-stage framework that separately trains instance feature extractors and feature aggregation networks. To enhance classification performance, researchers make contributions at both stages. For feature extractors, they introduce a variety of architectures from convolution neural networks to vision Transformers \cite{chen2022scaling, li2023survival}, and convert the training scheme from ImageNet pre-training to self-supervised learning \cite{dsmil, qu2022dgmil}. In the meantime, some researchers focus on feature aggregation and develop the non-parametric pooling to learnable ones \cite{clam, campanella2019clinical, transmil}. Among these methods, ABMIL \cite{abmil} attracts lots of attention for its characteristic of capturing discriminative instances. In this paper, we expand ABMIL with an attribute scoring mechanism and two attribute constraints, achieving accurate WSI classification and tumor localization.

\subsection{Parameter-efficient Fine-tuning}
Parameter-efficient fine-tuning techniques are first proposed in natural language processing because it is impractical to fully fine-tune the large language models for various downstream tasks \cite{houlsby2019parameter}. Their goal is to reach the performance of fully fine-tuning with low training costs (e.g., few training parameters and training time). Lately, parameter-efficient transfer learning has been developed in computer vision \cite{he2023parameter}, where adapter-based methods have yielded great success \cite{yang2023aim, chen2022conv}. Inspired by these works, AttriMIL incorporates adapters into the ImageNet pre-trained model, reducing the domain bias between natural and pathological images for better feature extraction.

\section{Methodology}
\label{sec:methodology}
\figref{fig:overview} shows the overview of AttriMIL. In this section, we first revisit MIL formulations and ABMIL. AttriMIL is then described in detail.

\subsection{Preliminaries}
\subsubsection{Formulate MIL}
In MIL, any input WSI is considered as a bag with multiple instances. Take binary classification as an example, let $X=\{(\mathbf{x}_1, y_1), ..., (\mathbf{x}_N, y_N)\}$ as a WSI bag, where $\mathbf{x}_i$ is the $i^{th}$ instance with an unavailable label $y_i \in \{0, 1\}$. Under the standard MIL assumption, the bag label $Y$ is formed as:
\begin{equation}
    Y = \left\{\begin{aligned}
    0, \quad & \text{iff}\ \sum\nolimits_{i} y_i = 0, \\
    1, \quad & \text{otherwise}.
    \end{aligned} \right.
\label{eq1}
\end{equation}
Generally, deep MIL yields the bag prediction through three steps \cite{lin2023interventional}. (1) Instance transformation: a feature extractor $f(\cdot)$ is used to extract instance-level feature $\mathbf{h}_i$, (2) instance aggregation: a pooling function $\sigma(\cdot)$ is targeted for bag feature $\mathbf{z}$, and (3) bag transformation: a bag-level classifier $g(\cdot)$ is applied for prediction. The above procedures can be formulated as:
\begin{equation}
\mathbf{h}_i = f(\mathbf{x}_i),\  \mathbf{z} = \sigma(\mathbf{h}_1, ..., \mathbf{h}_N),\  \hat{Y} = g(\mathbf{z}),
\label{eq2}
\end{equation}
where the $\sigma(\cdot)$ needs to be a permutation-invariant function to maintain the spatial invariance of MIL methods.

\subsubsection{Revisit ABMIL}
ABMIL \cite{abmil} generates a weight for each instance via a gated attention mechanism. Let $\{\mathbf{h}_1, ..., \mathbf{h}_N\}$ be a bag of $N$ instance features, ABMIL obtains the bag representation $\mathbf{z}$ through:
\begin{equation}
\mathbf{z} = \sum_{i=1}^{N}\ a_i\mathbf{h}_i \in \mathbb{R}^{1\times M},
\label{eq3}
\end{equation}
where $M$ is the dimension of vector $\mathbf{z}$ and $\mathbf{h}_i$; $a_i$ represents the attention score of the $i^{th}$ instance and is calculated by:
\begin{equation}
a_i = \frac{\text{exp}\{\mathbf{w}^{\text{T}}(\text{tanh}(\mathbf{V} \mathbf{h}^{\text{T}}_{i})) \odot \text{sigmoid}(\mathbf{U} \mathbf{h}^{\text{T}}_{i}) \}}{\sum_{j=1}^{N}{\text{exp}\{\mathbf{w}^{\text{T}}(\text{tanh}(\mathbf{V} \mathbf{h}^{\text{T}}_{j})) \odot \text{sigmoid}(\mathbf{U} \mathbf{h}^{\text{T}}_{j}) \}}},
\label{eq4}
\end{equation}
where $\mathbf{w} \in \mathbb{R}^{L\times 1}$, $\mathbf{V} \in \mathbb{R}^{L\times M}$, $\mathbf{U} \in \mathbb{R}^{L\times M}$ are parameters, $\mathbf{h}_i \in \mathbb{R}^{1 \times M}$ is the feature of the $i^{th}$ instance, ``$\text{T}$'' is a transpose operation, ``$\odot$'' is an element-wise multiplication, $\text{tanh}(\cdot)$ is a tanh function, and $\text{sigmoid}(\cdot)$ means a sigmoid non-linearity.

In ABMIL, the attention score $a_i$ reflects the importance of the $i^{th}$ instance in shaping the bag representation, thus providing interpretability to the classification result. Based on attention scores, numerous works \cite{clam, tourniaire2023ms, wang2023hard, shi2023structure} explore various ways to further strengthen MIL performance.

\begin{figure}[!t]
\centerline{\includegraphics[width=0.90\columnwidth]
{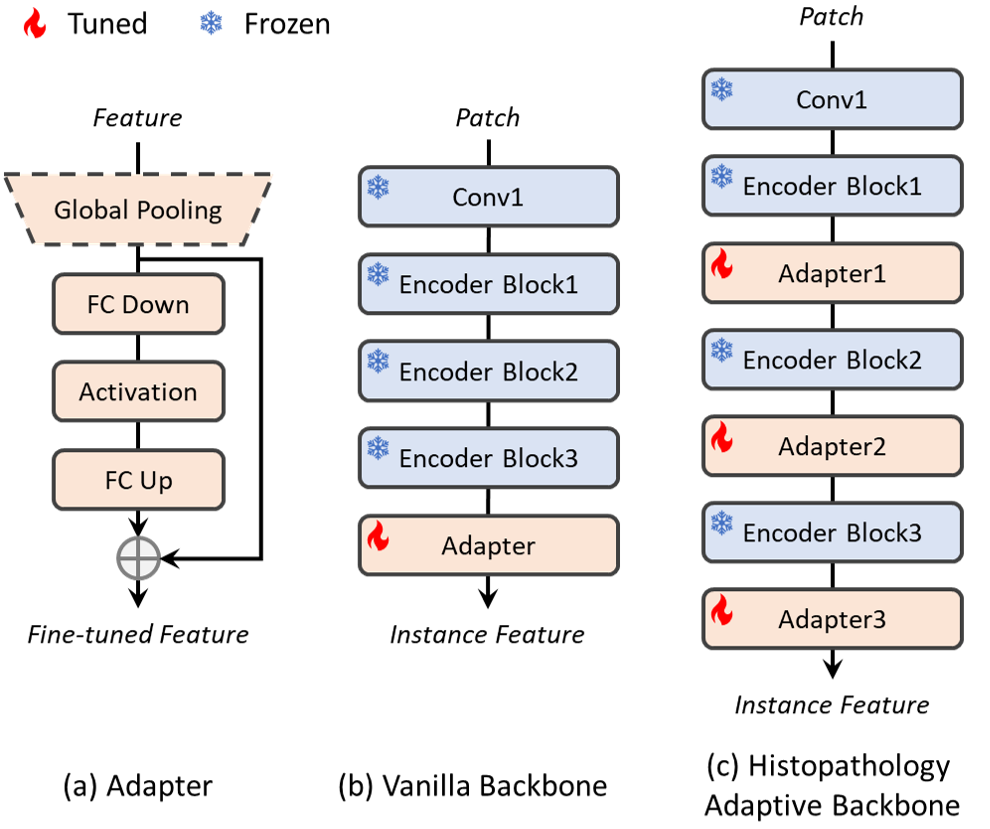}}
\caption{Illustration of histopathology adaptive backbone. In contrast to previous solutions, the histopathology adaptive backbone adopts feature adapters at different stages of the pre-trained network. For the adapter, the global pooling is used when training the current adapter.}
\label{fig:backbone}
\end{figure}

\subsection{Attribute Scoring Mechanism}
Although the attention score gives the importance of each instance in bag aggregation, it is not equivalent to the contribution to the final prediction. This difference may introduce noise in the selected instances when using attention for instance discrimination. To distinguish between instances more effectively, we dissect ABMIL and introduce an attribute scoring mechanism for precise measurement of instance attributes. 

Given a bag feature $\mathbf{z} \in \mathbb{R}^{1 \times M}$, ABMIL adopts a fully connected (FC) layer as a classification head for bag prediction:
\begin{equation}
\hat{Y} = b + \mathbf{z} \mathbf{c}\ ,
\label{eq5}
\end{equation}
where $b$ is a bias, $\mathbf{c} \in \mathbb{R}^{M\times 1}$ denotes the weight of the classification head. The network is inclined to classify the input as a positive bag when the value of $\hat{Y}$ is high, and as a negative bag when it is low. Combined with \eqtref{eq3}, \eqtref{eq5} can be further formed below:
\begin{equation}
\hat{Y} = b + \sum_{i=1}^{N} a_i \mathbf{h}_{i} \mathbf{c}\ ,
\label{eq6}
\end{equation}
where the second term reveals that the contribution of the $i^{th}$ instance to the final bag prediction is determined by the combined effect of the attention and the instance prediction rather than by the attention alone. Based on the above analysis, we define the attribute score $s_i$ of the $i^{th}$ instance as:
\begin{equation}
s_i = u_i \mathbf{h}_i \mathbf{c}\ ,
\label{eq7}
\end{equation}
where $u_i = \text{exp}\{\mathbf{w}^{\text{T}}(\text{tanh}(\mathbf{V} \mathbf{h}^{\text{T}}_{i})) \odot \text{sigmoid}(\mathbf{U} \mathbf{h}^{\text{T}}_{i}) \}$, denoting the unnormalized attention score of the $i^{th}$ instance. $u_i$ removes the impact of the instance number, enabling us to compare instance attributes across bags. $s_{i} \in (-\infty, +\infty)$ is the attribute score of the $i^{th}$ instance, in which the sign of the score indicates the instance attribute, while the absolute size of the value indicates the level of emphasis the network places on the instance. Essentially, AttriMIL converts ABMIL's sequential process of attention calculation and bag prediction into a parallel operation for attribute scoring. The bag prediction is obtained by the sum of normalized instance attribute scores (\textit{score aggregation} in AttriMIL) with a learnable bias as \eqtref{eq6}.

\subsection{Spatial Attribute Constraint}
Similar to birds of a feather flocking together, image tiles in a WSI exhibit significant spatial relationships, with patches of similar attributes often clustering together. By leveraging this prior information, we can alleviate errors in instance discrimination, thereby enhancing tumor localization and bag classification. With attribute scores, this paper proposes a simple but effective strategy, named spatial attribute constraint, to establish the spatial relation between patches.

Given an input WSI with $N$ instances, the spatial attribute constraint can be formulated as the sum of differences between adjacent instances:
\begin{equation}
\mathcal{L}_{spatial} = \frac{1}{N}\sum_{(i, j)} \sqrt{(s_{i, j} - s_{i+1, j})^{2} + (s_{i, j} - s_{i, j+1})^{2}},
\label{eq8}
\end{equation}
where $s_{i, j}$ denotes the attribute score of the instance located at $i, j$ position within a WSI. $s_{i+1, j}$ and $s_{i, j+1}$ represent the attribute scores of the instances below and to the right of the current instance, respectively.

\begin{figure*}[!t]
\centerline{\includegraphics[width=0.93\textwidth]
{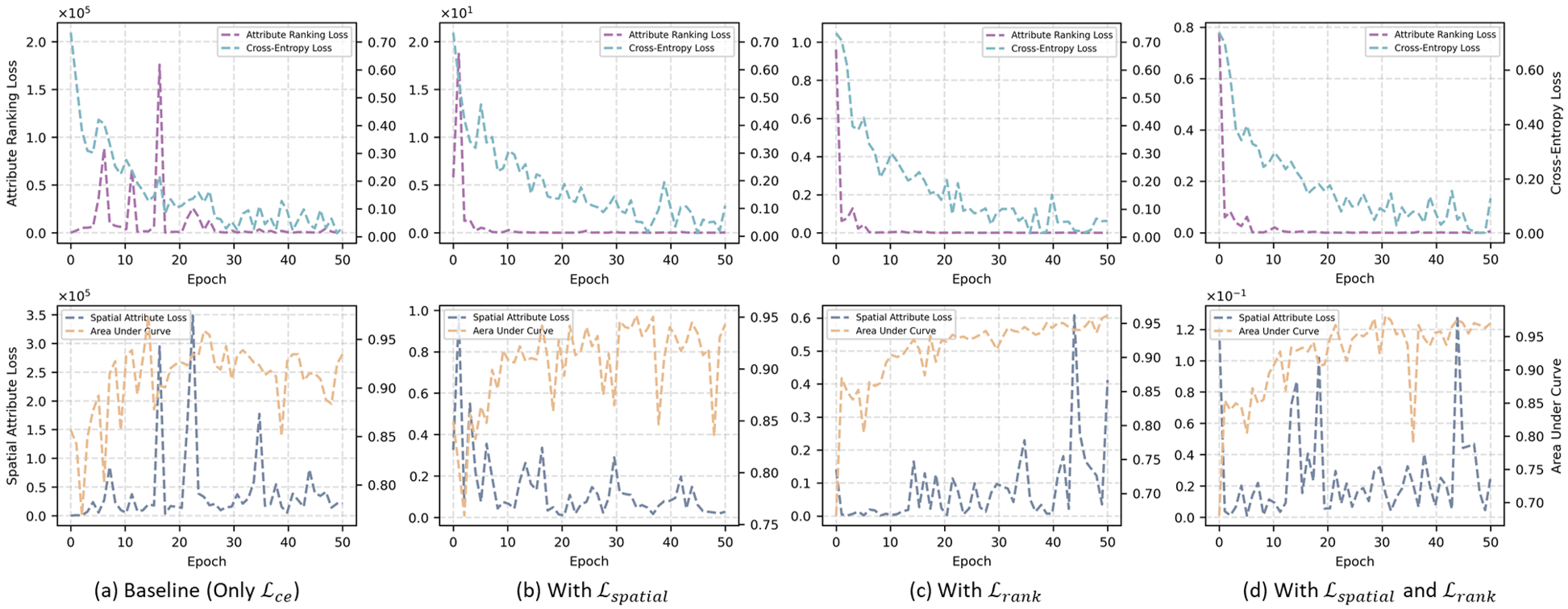}}
\caption{Training loss curves and AUC changes of the validation set under different loss constraints on the Camelyon16 dataset. $\alpha$ is set as 0.1 for the spatial attribute loss and $\beta$ is set to 0.001 for the attribute ranking loss.}
\label{fig:losscurves}
\end{figure*}

To implement the spatial attribute constraint efficiently, the coordinates and neighbors of each instance are recorded in pre-processing. For edge instances where adjacent positions are missing due to the irregularity of the segmented tissue, we consider their missing adjacent positions as their own, simplifying the computation. As shown in \figref{fig:overview}, for an input WSI, we calculate the average spatial attribute constraint of each branch to constrain AttriMIL in training.

\subsection{Attribute Ranking Constraint}
In addition to the intra-slide correlation, we also attempt to capture the instance correlation across WSIs, and thus enhance the model's instance attribute perception. Specifically, our goal is that positive instances exhibit higher attribute scores than negative ones, which can be formulated as:
\begin{equation}
s(Q^{p}) > s(Q^{n}),
\label{eq9}
\end{equation}
where $Q^{p}$ and $Q^{n}$ represent positive and negative instances, $s(Q^{p})$ and $s(Q^{n})$ represent the corresponding instance attribute scores respectively. The above ranking function is acceptable if the instance-level annotations are known during training. However, lacking instance-level annotations precludes the direct application of \eqtref{eq9}. Instead, the following multiple instance ranking function can be used:
\begin{equation}
\mathop{\max}\limits_{i\in X^p}s({Q}^{p}_{i}) > \mathop{\max}\limits_{i\in X^n}s({Q}^{n}_{i}),
\label{eq10}
\end{equation}
where $\max$ is taken over all patches in each bag. Different from the comparison in a single WSI, \eqtref{eq10} implements ranking only on the two instances having the highest attribute score respectively in the positive and negative bags. The patch with the highest attribute score likely represents a true positive instance, whereas the highest-score patch in a negative bag is considered a hard instance, closely resembling a positive instance yet being negative. Based on \eqtref{eq10}, we try to widen the gap in attribute scores between positive and negative instances, as follows:
\begin{equation}
\begin{aligned}
\mathcal{L}_{ra}&_{nk} = \max(0, -\mathop{\max}\limits_{i\in X^p}s(Q^{p}_{i}) + \mathop{\max}\limits_{i\in X^n}s(Q^{n}_{i})) \\
& + \max(0, -\mathop{\max}\limits_{i\in X^p}s(Q^{p}_{i}))
+ \max(0, \mathop{\max}\limits_{i\in X^n}s(Q^{n}_{i})),
\end{aligned}
\label{eq11}
\end{equation}
where the first item highlights the distance between instances of different attributes, the second and the third items use 0 as a threshold to constrain the attributes of the selected instances.

The attribute ranking loss can be extended for a subtype classification task in the multi-class attribute scoring mechanism. As shown in gray dotted boxes of \figref{fig:overview}, we adopt a positive bank and a negative bank for each subtype branch to record instances, where their capacity is set to $\text{K}$. For an input WSI in a branch, we use the highest attribute score instance to update the positive or negative bank according to whether the label of the WSI corresponds to the branch's subtype. Then, we select the top $\text{K}$ instances to calculate attribute rank loss with the recorded $\text{K}$ instances one by one and obtain the mean value, since a positive slide always contains multiple positive instances. The final attribute ranking loss is the average of ranking losses for each branch. In experiments, we set $\text{K}$ to 4 as the minimum positive instance number within a positive WSI is 4 in the training process.

\subsection{Histopathology Adaptive Backbone}
Within the bag-based MIL framework, the feature extractor is crucial for embedding patches into deep features to capture semantic information. To enhance instance representation, researchers \cite{dsmil, qu2022dgmil, chen2022scaling} investigate self-supervised learning for training a powerful extractor. However, their performance does not consistently outperform the ImageNet pre-trained backbone in pathological image classification tasks, as the final prediction is also determined by the aggregator \cite{bredell2023aggregation}. On the other hand, the pre-trained models have been trained on large-scale image datasets and have demonstrated excellent performance in various downstream tasks. We believe that they could provide pathological features through fine-tuning. In fact, this is a prevalent scheme in existing works \cite{transmil, clam}, where they generally integrate multi-layer perceptron (i.e., an adapter) after a pre-trained network to optimize instance-level features. In this paper, we develop a histopathology adaptive backbone, encouraging AttriMIL to perceive pathological information at various levels for better feature collection.

As shown in \figref{fig:backbone} (a), the adapter used in AttriMIL has a bottleneck architecture that consists of two fully connected (FC) layers and an activation layer in the middle. The first FC layer projects the input to a lower dimension and the second projects it to the original dimension. To enhance feature representation, we apply adapters after different stages of the feature extractor and introduce a progressive learning scheme for optimizing them. To be specific, we train the adapter progressively based on the depth of the network. Firstly, we use the first stage of the pre-trained model as the instance feature extractor and optimize the $1^{st}$ adapter. In the process, an \textit{auxiliary classifier} is used to perform score aggregation and add bias. The training loss $\mathcal{L}$ can be expressed as:
\begin{equation}
\mathcal{L} = \mathcal{L}_{ce} + \alpha \mathcal{L}_{spatial} + \beta \mathcal{L}_{rank},
\label{eq12}
\end{equation}
where $\mathcal{L}_{ce}=CE(Y, \hat{Y})$ is a cross-entropy loss for bag classification, $\alpha$ and $\beta$ are used to balance the three terms. 

With the trained adapter, we further collect the second stage output of the optimized pre-trained model, and then train the $2^{nd}$ adapter. By repeating the above steps, the training of histopathology adaptive backbone is completed.

\begin{figure*}[!t]
\centerline{\includegraphics[width=0.95\textwidth]
{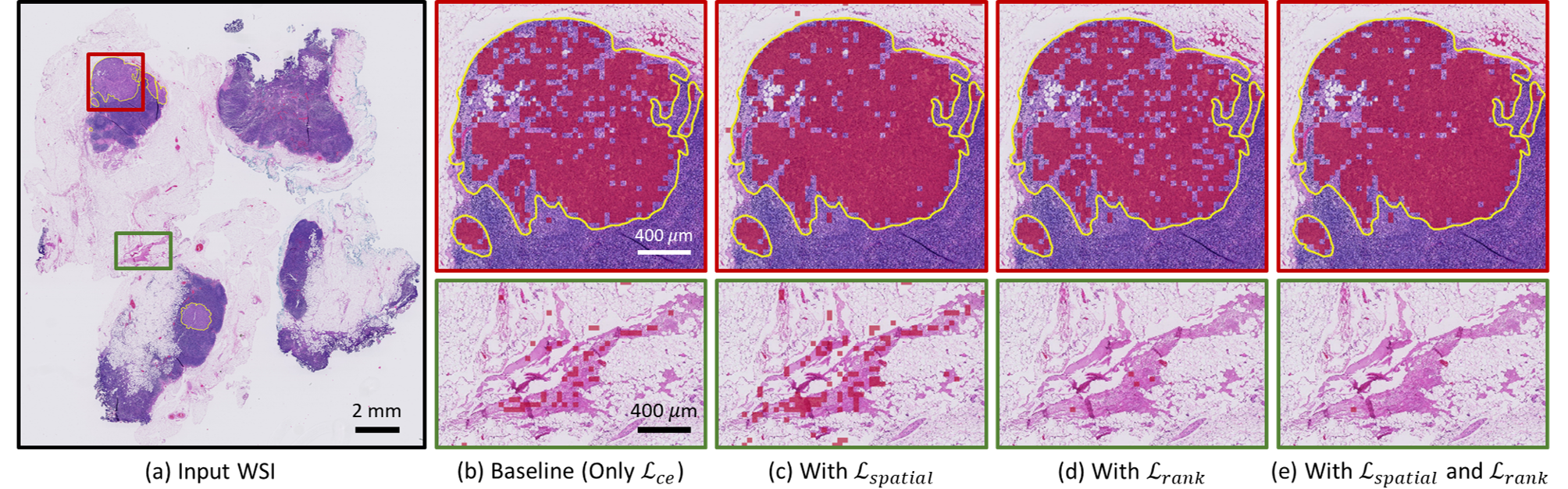}}
\caption{Visual analyses of the spatial attribute constraint and attribute ranking constraint. (a) presents a WSI from the Camelyon16 testing set, with tumor regions surrounded by yellow curves. (b)-(e) zoomed in area in the boxes of (a). In (b)-(e), tumor areas identified by the model (instance attribute score is greater than 0 in the tumor branch) are highlighted in red.}
\label{fig:lossablation}
\end{figure*}

\section{Experiments}
\label{sec:experiments}
In this section, extensive experiments are performed to validate the effectiveness of the proposed approaches. In Section \ref{sub:datasets}, we introduce the datasets and evaluation metrics used in our experiments. The implementation details are described in Section \ref{sub:detail}. Next, we present ablation studies and comparison results in Section \ref{sub:ablation} and \ref{sub:comparison}, respectively.

\begin{table}[t]
\vspace{-0.35cm}
\caption{Evaluation of spatial attribute constraint and attribute ranking constraint in terms of AUC (\%). The best result is shown in \textbf{bold}, and the second-best result is \underline{underlined}. P-values (baseline vs. best) are all less than 0.05.}
\centering
\resizebox{\columnwidth}{!}{%
\begin{tabular}{@{}c|c|cccccc@{}}
\toprule
Datasets                                         & \diagbox{$\alpha$}{$\beta$}  & $0$      & $0.001$     & $0.010$      & $0.100$       & $1.000$        & $10.000$   \\ \midrule
\multirow{6}{*}{Camelyon16}                      & $0$                          & $86.17$  & $88.52$     & $88.01$     & $88.27$     & $85.79$    & $84.21$ \\
                                                 & $0.001$                      & $86.56$  & $87.14$     & $87.22$     & $86.96$     & $84.77$    & $84.09$ \\
                                                 & $0.010$                      & $87.62$  & $88.15$     & $87.12$     & $\underline{89.46}$     & $83.18$    & $82.54$ \\
                                                 & $0.100$                      & $89.01$  & $\mathbf{91.31}$     & $89.26$     & $88.34$     & $86.42$    & $82.93$ \\
                                                 & $1.000$                      & $87.76$  & $88.16$     & $87.96$     & $88.97$     & $87.29$    & $83.72$ \\
                                                 & $10.000$                     & $86.76$  & $87.79$     & $87.19$     & $86.58$     & $86.86$    & $82.81$ \\ \midrule
\multirow{6}{*}{TCGA-NSCLC}                      & $0$                          & $94.36$  & $95.03$     & $94.97$     & $94.43$     & $94.43$    & $94.99$ \\
                                                 & $0.001$                      & $95.23$  & $95.13$     & $95.22$     & $95.46$     & $95.38$    & $94.24$ \\
                                                 & $0.010$                      & $95.23$  & $95.20$     & $94.95$     & $94.38$     & $\mathbf{95.49}$    & $94.51$ \\
                                                 & $0.100$                      & $95.30$  & $\underline{95.47}$     & $94.99$     & $94.89$     & $95.46$    & $94.29$ \\
                                                 & $1.000$                      & $95.32$  & $95.40$     & $94.72$     & $94.63$     & $95.43$    & $94.39$ \\
                                                 & $10.000$                     & $95.30$  & $95.43$     & $94.33$     & $94.46$     & $94.20$    & $94.11$ \\ \bottomrule
\end{tabular}}
\label{tab:parameters}
\vspace{-0.35cm}
\end{table}

\subsection{Datasets and Evaluation Metrics}
\label{sub:datasets}
\subsubsection{Dataset Description}
Our methods are evaluated on three public datasets: Camelyon16 \cite{camelyon16}, TCGA-NSCLC, and UniToPatho \cite{unitopatho}. \textbf{Camelyon16} \cite{camelyon16} is a breast cancer detection dataset that consists of 399 slides in 2 classes (normal and tumor). Following \cite{clam}, we remove the background and crop the WSI into 256$\times$256 sized non-overlapping patches at 20$\times$ magnification. We use the official split of 270/129 slides for training/testing and report the testing results. 
\textbf{TCGA-NSCLC} (\href{https://portal.gdc.cancer.gov}{https://portal.gdc.cancer.gov}) is a non-small cell lung cancer dataset, including 507 lung adenocarcinoma (LUDA) slides from 444 patients and 486 lung squamous cell carcinoma (LUSC) slides from 452 patients. Following \cite{clam}, we crop the foreground of each WSI into 256$\times$256 sized non-overlapping patches, obtaining approximately 15,000 patches per WSI. On the dataset, we conduct 4-fold cross-validation for evaluation. Specifically, we ensure that different slides from one patient case do not exist in both the training and testing sets, and then randomly split the data as the ratio of training: validation: testing = 60: 15: 25. 
\textbf{UniToPatho} \cite{unitopatho} is a colon cancer dataset comprising 9,536 hematoxylin and eosin (H\&E) stained patches extracted from 292 WSIs. There are six classes in the dataset, namely, normal tissue (NORM), hyperplastic polyp (HP), tubular adenoma \& high-grade dysplasia (TA.HG), tubular adenoma \& low-grade dysplasia (TA.LG), tubulo-billous adenoma \& high-grade dysplasia (TVA.HG), and tubulo-villous adenoma \& low-grade dysplasia (TVA.LG). Based on the official code, we crop the provided images into 224$\times$224 sized patches without overlapping. In experiments, the official split of 204/88 slides is used for training/testing.

\subsubsection{Evaluation Metrics}
We evaluate the model performance using class-wised average accuracy (ACC), F$_1$-Score, and the area under the curve (AUC) score. The higher the value of these indicators, the better the performance of the method. In experiments, we took AUC as the major evaluation metric and conducted a Delong test on it to verify the statistical significance of improvements. To avoid randomness, we ran the experiments four times and reported the average metrics on Camelyon16 and UniToPatho datasets. For TCGA-NSCLC, we reported average metrics from 4-fold cross-validation.

\begin{table}[t]
\vspace{-0.35cm}
\caption{Evaluation of the use of adapters in histopathology adaptive backbone. P-values (adapter3 vs. adapter1 + adapter2 + adapter3) are all less than 0.05.}
\centering
\resizebox{\columnwidth}{!}{%
\begin{tabular}{@{}ccc|cc|cc@{}}
\toprule
\multicolumn{3}{c|}{ImageNet Pre-trained} & \multicolumn{2}{c|}{Camelyon16} & \multicolumn{2}{c}{TCGA-NSCLC} \\ \midrule
Adapter1    & Adapter2    & Adapter3   & ACC ($\%$)            & AUC ($\%$)           & ACC ($\%$)            & AUC ($\%$)           \\ \midrule
\checkmark            &              &            & $65.41$          & $72.31$          & $80.52$          & $86.32$         \\
             & \checkmark            &            & $79.84$          & $83.38$          & $85.44$          & $91.23$         \\
             &              & \checkmark          & $84.49$          & $86.17$          & $88.45$          & $94.36$         \\
\checkmark            & \checkmark            &            & $81.21$          & $84.52$          & $86.57$          & $91.45$         \\
\checkmark            &              & \checkmark          & $85.38$          & $89.24$          & $86.60$          & $94.58$         \\
             & \checkmark            & \checkmark          & $85.72$          & $89.92$          & $88.92$          & $94.77$         \\
\checkmark            & \checkmark            & \checkmark          & \underline{$86.63$}          & $\mathbf{91.12}$          & $\mathbf{89.50}$          & $\mathbf{95.11}$         \\ \midrule
\multicolumn{3}{c|}{SimCLR \cite{simclr}}   & $\mathbf{86.87}$          & \underline{$89.95$}          & \underline{$89.41$}          & \underline{$95.03$}         \\
\multicolumn{3}{c|}{MoCov2 \cite{moco_v2}}     & $86.21$          & $88.67$          & $89.05$         & $94.88$         \\ \bottomrule
\end{tabular}}
\label{tab:adapters}
\vspace{-0.35cm}
\end{table}

\begin{table*}[!t]
\caption{Quantitative comparison of our AttriMIL and state-of-the-art methods. The best performance is highlighted in \textbf{bold}, and the second-best is \underline{underlined}. P-values (ABMIL vs. AttriMIL) are all less than 0.05.}
\centering
\resizebox{\textwidth}{!}{%
\begin{tabular}{c|ccc|ccc|ccc}
\toprule
\multirow{2}{*}{Methods} & \multicolumn{3}{c|}{Camelyon16} & \multicolumn{3}{c|}{TCGA-NSCLC} & \multicolumn{3}{c}{UniToPatho}                                   \\ \cmidrule{2-4} \cmidrule{5-7} \cmidrule{8-10} 
                         & ACC (\%)  & F$_1$-Score (\%)  & AUC (\%)  & ACC (\%)  & F$_1$-Score (\%)  & AUC (\%)  
                         & ACC (\%)  & F$_1$-Score (\%)  & AUC (\%) \\ \midrule
Mean-Pooling             & $79.56\pm3.15$    & $70.16\pm7.59$      & $80.01\pm7.25$    & $84.78\pm1.26$   & $84.74\pm1.36$    & $92.84\pm1.14$   & $66.76\pm3.34$   & $\underline{52.12}\pm2.68$      & $88.25\pm1.27$                  \\
Max-Pooling              & $81.32\pm6.63$    & $72.24\pm4.46$      & $84.65\pm5.03$   & $85.65\pm1.50$   & $83.97\pm2.22$    & $91.16\pm0.79$  & $53.69\pm2.18$    & $35.28\pm4.99$      & $73.68\pm1.84$                      \\
MIL-RNN \cite{campanella2019clinical} & $83.72\pm2.95$    & $76.40\pm3.87$      & $83.21\pm3.65$   & $86.15\pm2.31$ 
 & $84.89\pm3.55$    & $92.25\pm1.03$  & $51.14\pm4.18$    & $43.15\pm2.78$      & $78.18\pm0.62$                      \\
DSMIL  \cite{dsmil}                 & $\underline{88.75}\pm2.21$    & $\underline{84.33}\pm2.98$      & $\underline{92.33}\pm1.43$      & $\underline{89.50}\pm2.15$  & $\mathbf{92.61}\pm2.93$    & $\underline{95.52}\pm1.21$   & $\mathbf{67.04}\pm2.43$    & $49.69\pm3.55$      & $\underline{88.48}\pm1.00$ \\
CLAM-SB \cite{clam}                 & $83.72\pm1.12$    & $77.89\pm1.49$      & $90.66\pm2.78$   & $88.60\pm2.47$  & $88.61\pm2.19$    & $95.39\pm1.45$   & $52.27\pm3.40$    & $41.61\pm1.21$      & $84.79\pm1.06$                    \\
CLAM-MB \cite{clam}                 & $84.41\pm1.33$    & $78.06\pm2.05$      & $90.05\pm2.35$   & $88.90\pm2.57$    & $89.10\pm2.31$    & $95.02\pm1.53$    & $58.23\pm2.58$    & $46.86\pm2.40$      & $84.27\pm0.70$                    \\
DGMIL \cite{qu2022dgmil}   & $82.49\pm2.93$    & $75.10\pm2.45$      & $88.86\pm3.10$  & $88.84\pm1.43$ 
        & $88.71\pm1.57$    & $94.55\pm1.29$   & -        & -          & -                            \\
TransMIL \cite{transmil}                 & $83.72\pm2.33$    & $81.50\pm3.56$      & $88.86\pm2.85$  & $88.46\pm2.61$                     & $88.42\pm3.74$    & $94.32\pm2.01$    & $63.92\pm4.10$    & $42.73\pm6.67$      & $87.02\pm1.98$                     \\
DTFD-MIL \cite{zhang2022dtfd}                & $86.61\pm0.88$    & $79.52\pm1.07$      & $88.82\pm2.49$   & $88.46\pm2.01$   & $86.90\pm2.26$    & $93.77\pm1.24$  & $64.53\pm3.92$    & $43.75\pm5.24$      & $86.45\pm1.92$                  \\ 
PMIL \cite{PMIL}                & $87.86\pm1.59$    & $80.15\pm2.16$      & $90.20\pm2.40$  & $88.54\pm1.92$   & $86.77\pm2.42$    & $94.83\pm1.69$      & -    & -      & -                   \\ \midrule
ABMIL \cite{abmil}  & $84.49\pm1.65$    & $76.74\pm3.06$      & $88.75\pm3.15$     & $88.90\pm0.51$  & $88.59\pm0.79$    & $94.95\pm0.30$ & $57.38\pm1.71$    & $44.27\pm4.45$      & $85.37\pm0.37$  \\
AttriMIL (Ours)            & $\mathbf{90.69}\pm1.02$    & $\mathbf{87.23}\pm1.78$      & $\mathbf{93.90}\pm1.23$     & $\mathbf{90.38}\pm2.32$     & $\underline{90.24}\pm2.21$    & $\mathbf{96.13}\pm1.20$ & $\underline{66.92}\pm3.41$    & $\mathbf{55.97}\pm4.71$      & $\mathbf{88.99}\pm1.31$ \\ \bottomrule
\end{tabular}}
\label{tab:comparison}
\vspace{-0.35cm}
\end{table*}

\subsection{Implementation Details}
\label{sub:detail}
We employ an Adam optimizer with a constant learning rate of 2e-4 for updating learnable weights during the training phase. The mini-batch size for training is set to 1 (bag). The hyper-parameters $\alpha$ and $\beta$ of the loss function (\eqtref{eq12}) are set as 0.1 and 0.001 in training each adapter by default. We adopt the first three blocks of an ImageNet pre-trained ResNet-50 as the feature extractor's backbone. Notably, we replace batch normalization with group normalization \cite{wu2018group} in the backbone to avoid interactions between instances during feature extraction. All experiments are implemented on the PyTorch 1.10.0 using an Nvidia GeForce RTX 3090 GPU.

\subsection{Ablation Study}
\label{sub:ablation}
In this subsection, we explore and analyze the effectiveness of each component. Details are described in following parts.

\begin{figure*}[!t]
\centerline{\includegraphics[width=0.95\textwidth]
{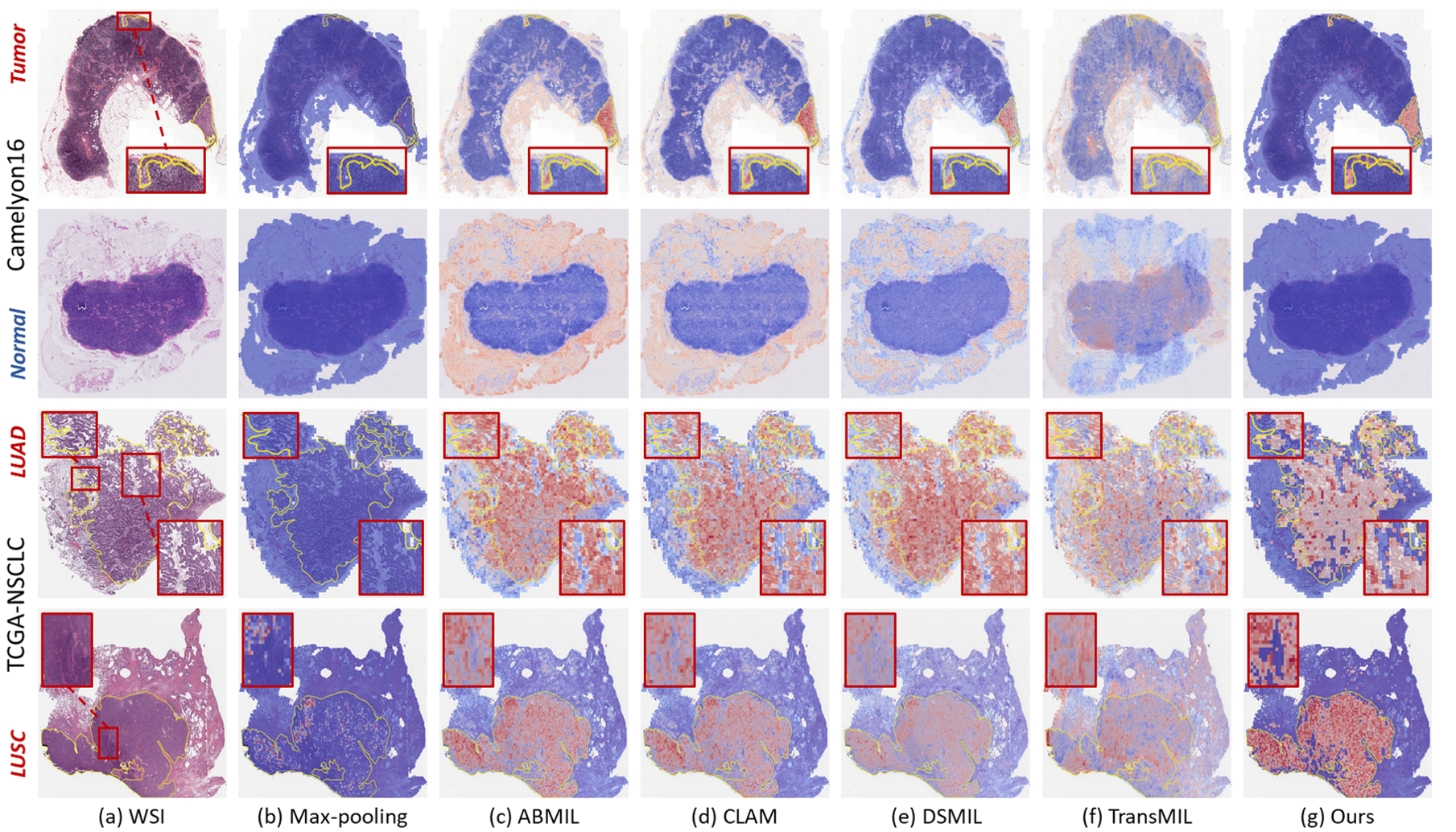}}
\caption{Qualitative comparison of our AttriMIL and state-of-the-art methods. Tumor regions in each WSI are surrounded by yellow curves, and red boxes highlight the salient differences of each method in tumor localization. For (b), instance prediction results are used for patch discrimination. For (c)-(f), attention scores are re-scaled from min-max to [0, 1]. For (g), attribute scores are used for instance discrimination. According to the meaning of attribute score signs, score values less than 0 are set as 0, and values greater than 0 are re-scaled from min-max to [0.5, 1].}
\label{fig:Qualitative}
\end{figure*}

\subsubsection{Evaluation of Spatial Attribute Constraint}
In this part, we investigate the impact of the spatial attribute constraint on WSI classification. We first conducted a hyper-parameter optimization experiment on the Camelyon16 and TCGA-NSCLC datasets, with the results detailed in \tabref{tab:parameters}. From the column of $\beta=0$, we can observe two points. (1) Regardless of the value of $\alpha$ ($>$0), the incorporation of spatial attribute constraint consistently increases the performance of the baseline on both datasets, indicating the importance of patch spatial correlation in WSI classification. (2) The effect of spatial attribute constraint is related to the dataset, as the optimal value of $\alpha$ is inconsistent in different datasets. We argue the phenomenon is attributed to the differences in instance distribution across the datasets. Compared with TCGA-NSCLC, the tumor area ratio is too small on Camelyon16, which may cause overfitting for the negative instance when the value of $\alpha$ is large, thus yielding limited improvements.

To further demonstrate the effectiveness of spatial attribute constraint, we present the training curves in \figref{fig:losscurves}. Comparing (a) and (b), we observe a large spatial attribute loss (exceeding 350,000) in the vanilla training process, due to the baseline (ABMIL) not taking the spatial correlation of instances into account. Instead, the use of spatial attribute constraint enables the network to be aware of the intra-slide correlation and improves the bag classification performance. Additionally, \figref{fig:lossablation} intuitively shows the impact of spatial attribute constraint. As depicted in the upper box of (c) , the spatial constraint leads to fewer holes in the positive regions, underscoring the utility of the spatial attribute constraint in enhancing tumor localization. Nevertheless, establishing only spatial relationships within a single WSI is powerless to solve the problem of identifying hard samples, as shown in the bottom box of (c).

\subsubsection{Evaluation of Attribute Ranking Constraint}
The attribute ranking constraint is designed to enhance the model's learning capability by leveraging inter-slide correlations. \tabref{tab:parameters} presents the influence on the model performance under different settings of $\beta$. From the rows of $\alpha=0$, we obtain two observations: (1) When $\beta$ is set as 0.001, the performance of the baseline is improved by 2.35\% and 0.67\% on Camelyon16 and TCGA-NSCLC respectively, indicating that the attribute ranking constraint is beneficial to WSI classification. (2) A large $\beta$ value reduces the model performance on Camelyon16, likely because the large ranking loss makes the network overly focus on differences between partial instances and unreasonably decrease the overall loss.

\figref{fig:losscurves} (a) and (c) reflect the difference in the learning process when attribute rank loss is introduced, from which we can observe: (1) In the case of using only cross-entropy loss, the attribute ranking loss presents multiple peaks in the training process, and gradually approaches a low level (about 2,000) at the end of training. In other words, the baseline attempts to reduce the ranking loss in the training, which proves the rationality of the introduction of attribute ranking loss. (2) As shown in \figref{fig:losscurves} (c), the introduction of attribute ranking loss makes AUC stably rise in the training process, which indicates attribute ranking constraint is effective for facilitating MIL. In addition, \figref{fig:lossablation} intuitively shows the superiority of ranking loss in differentiating hard instances (bottom of (b) and (d)). The model trained with $\mathcal{L}_{rank}$ significantly reduces the wrong classification of instances compared to the baseline.

Furthermore, we explore the joint use of the spatial attribute constraint and attribute ranking constraint. As listed in the \tabref{tab:parameters}, the combination of both constraints significantly improves the performance of baseline (P-values$<$0.05), boosting the AUC of baseline by 5.14\% and 1.10\% on Camelyon16 and TCGA-NSCLC, respectively. The results in \figref{fig:lossablation} (e) visually illustrate the advantages of the dual constraints. In comparison to (b), (c), and (d), the model with constraints performs better in localizing the tumor regions and successfully distinguishes hard negative instances. These findings show that attribute constraints can synergistically contribute to model training, and the introduction of intra-slide and inter-slide correlations is meaningful for WSI classification. Moreover, these significant improvements demonstrate the advance of the attribute scores, since the implementation of both loss functions relies on the attribute scoring mechanism.

\subsubsection{Evaluation of Histopathology Adaptive Backbone}
In AttriMIL, histopathology adaptive backbone is used to obtain the optimized instance features. In ablation experiments, we adopted an ImageNet pre-trained ResNet-50 as a backbone, inserted adapters, and successively trained them using the cross-entropy loss function. \tabref{tab:adapters} shows the impact of these adapters on WSI classification. From the table, we can derive the following conclusions. (1) The shallow features extracted by the feature extractor are able to reflect pathological property to a certain extent, as the ResNet-50 with adapter1 achieves 72.31\% AUC on Camelyon16 and 86.32\% AUC on TCGA-NSCLC. (2) Multiple adapters facilitate the instance feature extraction, and our histopathology adaptive backbone (adapter1 + adapter2 + adapter3) achieves the best performance compared to other settings. In addition, following \cite{dsmil} and \cite{Kang_2023_CVPR}, we present the classification performance of contrastive learning methods (i.e., SimCLR \cite{simclr} and MoCov2 \cite{moco_v2}) in \tabref{tab:adapters}. In comparison with these methods, our method achieves superior AUC results. This is attributed to the property that our approach focuses on pathological image classification, enabling the backbone to adaptively refine features at various levels for powerful task-specific representation.

\subsection{Comparisons with State-of-the-Art Methods}
\label{sub:comparison}
In this part, we present classification results for both binary classification and multiple classification. Binary classification tasks contain positive/negative classification over Camelyon16 and LUSC/LUAD subtype classification over TCGA-NSCLC. The multiple classification task refers to NORM, HP, TA.HG, TA.LG, TVA.HG, and TVA.LG classification over UniToPatho. Table \ref{tab:comparison} lists the quantitative comparisons of our AttriMIL and current state-of-the-art methods, where Mean-pooling and Max-pooling are utilized to highlight the superiority of each method. For a fair comparison, we obtain the experimental results by rerunning their released code with their published instance features.

\subsubsection{Quantitative Comparison} In Camelyon16, the tumor areas only occupy a small proportion of each positive WSI (the average proportion is less than 10\%). Attention-based methods like ABMIL \cite{abmil}, CLAM \cite{clam}, DSMIL \cite{dsmil}, and TransMIL \cite{transmil} consistently outperform the traditional Mean-pooling and Max-pooling. In terms of AUC, ABMIL, DTFD-MIL \cite{zhang2022dtfd} and CLAM are at least 3\% lower than AttriMIL, as these methods ignore the correlation between patches in WSI classification. TransMIL and DSMIL only model the relations between instances within a single WSI, resulting in limited performance. Notice that, PMIL \cite{PMIL} is a prototype-based MIL framework which considers the relations across WSIs. However, PMIL mainly emphasizes phenotypic differences between instances and neglects the pathology attribute, leading to inferior performance compared to AttriMIL. In TCGA-NSCLC, the positive WSI generally contains large tumor regions, consequently, all the methods perform better than on the Camelyon16 dataset. In comparison, AttriMIL outperforms other methods, with an increase in ACC and AUC of 0.88\% and 0.61\%, respectively. In the UnitoPatho dataset, as DGMIL and PMIL do not consider the multi-subtype classification tasks, Table \ref{tab:comparison} does not list their results. UniToPatho has an unbalanced distribution of subtypes and positive areas. In the dataset, our AttriMIL achieves advanced performance in terms of ACC, F$_1$-Score, and AUC, indicating that AttriMIL can be applied to multi-class problems with unbalanced data.

\begin{figure}[!t]
\centerline{\includegraphics[width=\columnwidth]
{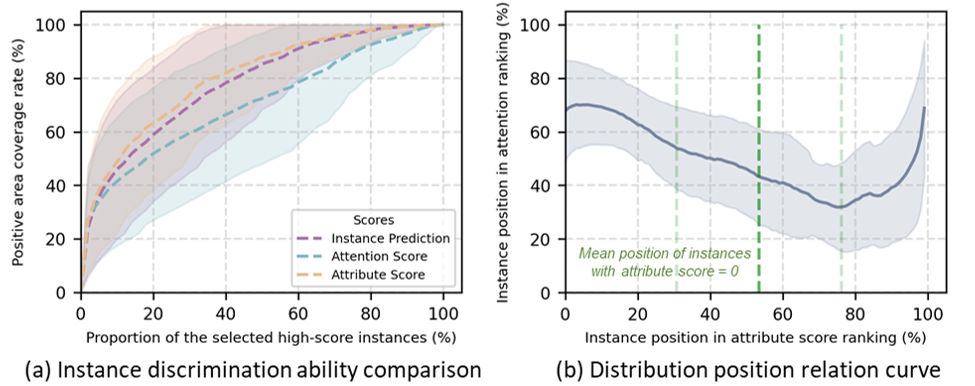}}
\caption{Instance discrimination ability comparison and distribution position relation between attention and attribute scores on the Camelyon16 dataset. In (a), the positive area coverage rate indicates the proportion of the number of positive instances within the high-score instances to the total number of positive instances. For (b), we rank instances based on attention and attribute scores and present the position relation between them, where 100\% represents the highest-score instance. The green lines (mean $\pm$ standard deviation) represent the ranking position where the instance's attribute score is equal to 0.}
\label{fig:relation}
\end{figure}

\begin{figure}[!b]
\centerline{\includegraphics[width=\columnwidth]
{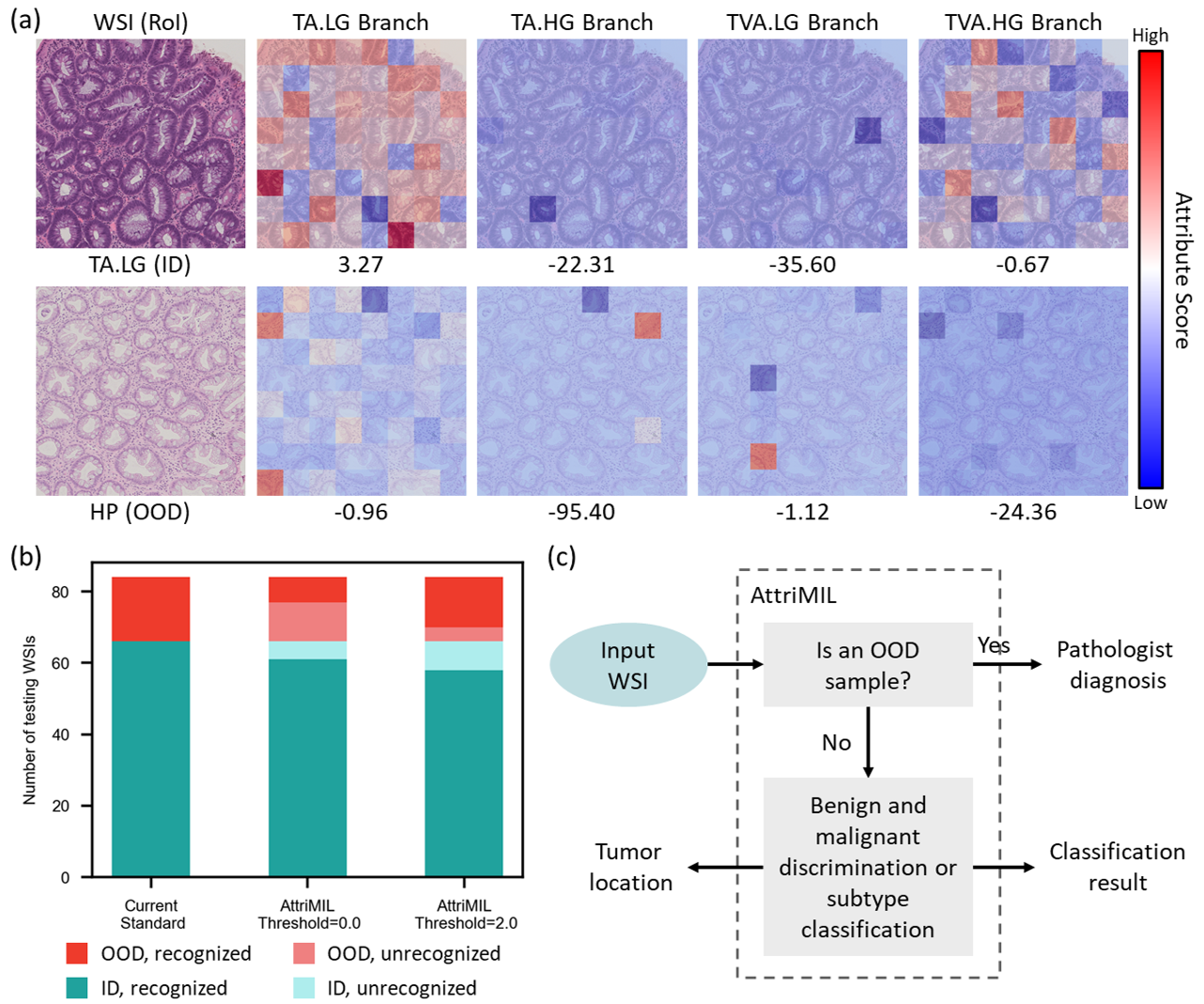}}
\caption{Out-of-distribution detection and potential applications using AttriMIL. (a) presents instance attribute score distribution (without normalized attention) and the aggregated bag score in the WSI's region of interest (RoI) on UniToPatho. For the OOD sample, the bag attribute score in each branch is less than 0. Bar plots in (b) show the OOD performance of AttriMIL under different thresholds. When the threshold is set to 2, AttriMIL can identify more OOD samples. (c) shows the potential impact of AttriMIL on computer-assisted pathological diagnosis.}
\label{fig:ood}
\end{figure}

\subsubsection{Qualitative Comparison} \figref{fig:Qualitative} intuitively illustrates the tumor localization capability of different methods, from which we can obtain three observations. (1) Compared to existing methods, our AttriMIL exhibits strong tumor localization power. As shown in the $1^{st}$ row, our method accurately captures tumors with extremely small areas. (2) Previous MIL frameworks always misidentify negative instances. By contrast, our method effectively distinguish the negative instances in various scenes (shown in the $1^{st}$ and $2^{nd}$ rows). This is due to the introduction of attribute scoring and attribute ranking loss, which offer an attribute measurement and a threshold ($0$ in this paper) for differentiating instance attributes. To a certain extent, attribute ranking loss introduces the advantages of Max-pooling into the attention-based MIL framework for locating discriminative instance. (3) In comparison to other methods, AttriMIL effectively identifies negative regions embedded in the tumor area. For the LUAD slide, AttriMIL assigns low attribute scores to the pulmonary acinus embedded in the tumor area. For the LUSC slide, AttriMIL successfully recognizes the connective tissues, including fibers and immune cells. These underscore AttriMIL's advancements in providing highly interpretable and accurate results for WSI analysis.

\section{Discussion and Conclusion}
\label{sec:disandcon}
\subsection{Relationship between Attention and Attribute Scores}
In this part, we explore the relationship between attention and attribute scores. We trained an ABMIL using the cross-entropy loss function on Camelyon16 and present comparisons of various scores in \figref{fig:relation}. \figref{fig:relation} (a) reports the instance discrimination ability of attention, attribute scores, and instance predictions, from which we can see that the attribute score outperforms the others in localizing positive regions while the attention performance is poor. \figref{fig:relation} (b) shows the relation between attention and attribute scores, where the distribution position relation curve is U-shape. This suggests that attention tends to focus on significantly positive and negative areas, and the attribute scoring mechanism improves attention for precise instance discrimination. Furthermore, from \eqtref{eq4} and \eqtref{eq6}, we can realize that the attention mechanism can be regarded as a special bag classifier, which weights instance in exponential form compared to the vanilla fully connected layer, enabling ABMIL to quickly converge. This is the reason why ABMIL has performed well in the instance attribute viewpoint.

\subsection{AttriMIL for Out-of-Distribution Detection}
Different from previous frameworks, AttriMIL sums the attribute score of each instance for bag aggregation, where the aggregated score in each branch is meaningful due to the characteristic of attribute scores. Specifically, AttriMIL uses 0 as the threshold (refer to \eqtref{eq11}). If the bag score exceeds 0, it means that the input WSI corresponds to the attribute of the branch; otherwise, it does not belong to this branch. This feature allows AttriMIL to detect out-of-distribution (OOD) samples, i.e., when the bag scores of all branches are less than 0, the sample is considered to be an OOD bag. To verify this point, we conducted experiments on UniToPatho, where the in-distribution (ID) samples of TA.LG, TA.HG, TVA.LG, and TVA.HG were used for training, and the Norm and HP WSIs were considered OOD samples. \figref{fig:ood} visually shows the performance of AttriMIL in OOD detection, and thus our method provides a solution to establish a complete computer-assisted pathological diagnosis system.

\subsection{Conclusion}
In this paper, we propose a novel multiple instance learning (MIL) framework named Attribute-Driven MIL (AttriMIL) for whole-slide pathological image analysis. Unlike previous solutions, AttriMIL introduces an attribute scoring scheme that effectively measures the contribution of each instance to bag prediction, thereby quantifying the instance's attributes. Based on the quantification of instance attributes, two attribute constraints, namely spatial attribute constraint and attribute ranking loss, are developed to model the intra-slide and inter-slide correlations among instances, respectively. These correlations enhance the instance discrimination capability of the network and enable AttriMIL to achieve accurate tumor localization. Moreover, a histopathology adaptive backbone is applied in AttriMIL, which takes full advantage of the pre-trained model for improving instance feature representation. Extensive experiments on three benchmarks demonstrate the superiority of our method. Furthermore, AttriMIL shows potential in processing out-of-detection samples, providing a promising solution for building a complete pathological diagnosis system.
\bibliographystyle{IEEEtran}
\bibliography{AttriMIL.bib}

\end{document}